# RO-BENCH: LARGE-SCALE ROBUSTNESS EVALUATION OF MLLMS WITH TEXT-DRIVEN COUNTERFACTUAL VIDEOS

*Zixi Yang*[⋆†]   *Jiapeng Li*[⋆†]   *Muxi Diao*[⋆†]   *Yinuo Jing*[⋆†]   *Kongming Liang*[⋆‡]

⋆ Beijing University of Posts and Telecommunications

## ABSTRACT

Recently, Multi-modal Large Language Models (MLLMs) have demonstrated significant performance across various video understanding tasks. However, their robustness, particularly when faced with manipulated video content, remains largely unexplored. In this paper, we introduce Ro-Bench, the first benchmark for evaluating MLLMs on dynamic out-of-distribution (OOD) counterfactual video test sets. Ro-Bench incorporates high-quality, diverse and temporally relevant video data, by editing Style, Object, Background and their compositions. We evaluated eight recent video MLLMs and found that current models exhibit substantial performance degradation on Ro-Bench when exposed to counterfactual video content. Furthermore, we demonstrate that fine-tuning MLLMs with counterfactual data enhances robustness, achieving a 21.73% performance increase on Ro-Bench and a 12.78% improvement across 20 tasks in the MVBench dataset. These findings underscore the effectiveness of counterfactual data in enhancing the video understanding ability of MLLMs. The code and data will be released shortly.

*Index Terms*— Multi-modal Large Language Models, Robustness, Evaluation, Counterfactual

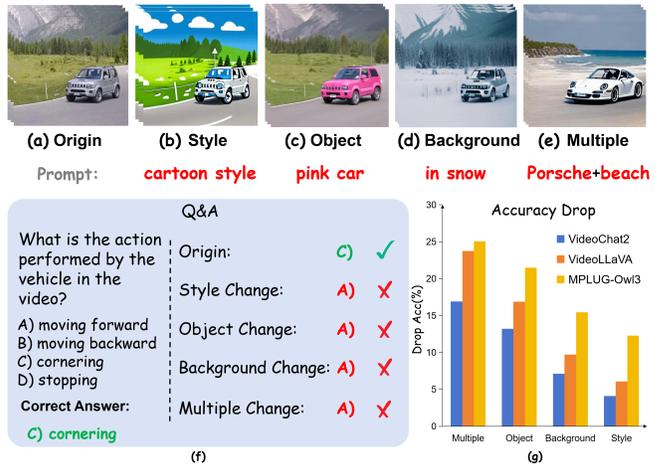

**Fig. 1**. Overview of the Ro-Bench benchmark. (a)-(e): Editing factors for automatic test video generation. (f): Example QA pairs. (g): MLLM performance drop, highlighting robustness to visual variations.

## 1. INTRODUCTION

Multi-modal Large Language Models (MLLMs) have demonstrated impressive performance across a wide range of tasks that require understanding both visual and textual inputs [1, 2, 3, 4, 5, 6, 7, 8]. As MLLMs are increasingly deployed in high-stakes domains, such as video-based content moderation, autonomous driving, and real-time surveillance, ensuring their robustness becomes a critical concern [9]. While these models perform well in controlled environments, their ability to maintain performance when faced with altered or manipulated inputs remains largely unexplored.

In recent years, there has been growing interest in developing out-of-distribution (OOD) benchmarks [10] to better evaluate model robustness. For instance, the LANCE [11] has demonstrated the value of using counterfactual image generation to assess model performance. However, these efforts have been largely focused on static images, leaving a gap in the evaluation of video understanding models. Moreover, current robustness benchmarks in the video domain often focus solely on noise or corruption-based testing, overlooking the rich temporal dynamics and complex attribute relationships present in real-world videos.

In this work, we are particularly interested in two research questions for MLLMs' robustness evaluation:

*RQ1: How do MLLMs perform on counterfactual videos, and what specific challenges do they face in understanding edited video content?*

*RQ2: How does the use of counterfactual videos influence MLLMs' performance, and can it enhance their ability to understand and interpret complex video content?*

To evaluate the robustness of video MLLMs, we introduce Ro-Bench, the first benchmark that systematically assesses model performance on dynamic OOD counterfactual video test sets. Ro-Bench features high quality, high diversity, temporal relevance, and diverse tasks. Our pipeline automatically generates realistic counterfactual videos through text-driven video editing methods and contribute 8.6k multiple-choice QA pairs across 4 video understanding tasks. In response to RQ1 and RQ2, we evaluate mainstream video MLLMs on Ro-Bench and introduce the **LLaVA-Next$_{Ro}$**, fine-tuned with counterfactual data, showed a substantial 21.73% boost in Ro-Bench compared to the baseline LLaVA-Next. Figure 1 provides a high-level overview of our editing factors, benchmark components, and key findings.

---

[1]† These authors contributed equally.
[2]‡ Corresponding Author: liangkongming@bupt.edu.cn

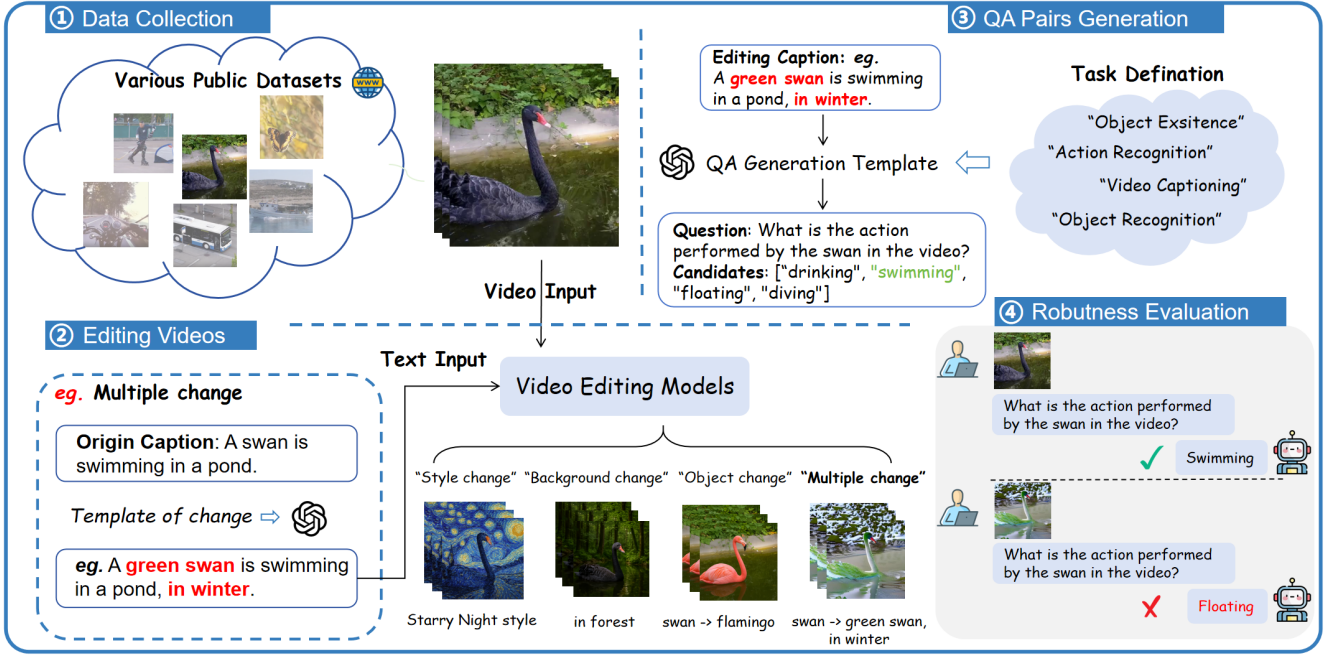

**Fig. 2. Ro-Bench Pipeline.** (1) Collects high-quality video from the Internet and open-source datasets, (2) uses a text-driven approach to generate counterfactual videos, (3) generates QA pairs based on different tasks, (4) evaluates model robustness.

To summarize, our contributions are as follows:

- We introduce Ro-Bench, the first benchmark that generates counterfactual video test sets for evaluating the robustness of video MLLMs.

- We introduce four innovative evaluation metrics to assess how textual prompts and original videos influence editing outcomes, helping to ensure high quality data

- Using Ro-Bench, we conduct a comprehensive evaluation of mainstream video MLLMs, revealing their insufficient robustness in understanding videos.

- We conduct experiments showing that training on counterfactual data improves performance on Ro-Bench and general performance on other benchmark tasks.

## 2. RO-BENCH

### 2.1. Data Source

Diverse data sources are essential for achieving sample diversity in Ro-Bench. We manually collect raw video and caption data from two main sources: publicly available datasets (DAVIS[12], TGVE[13], MSR-VTT[14], and BalanceCC[15]) and the Internet. The dataset includes four agent types: *Human*, *Animal*, *Landscape*, and *Object*, as well as four task types: *Action Recognition (AR)*, *Object Recognition (OR)*, *Object Existence (OE)*, and *Video Captioning (VC)*. The video sources are well-balanced, offering a diverse content across different agent categories and scenarios.

### 2.2. Data Processing Pipeline

We use text-driven video editing models to create new videos that reflect the changes in the text. We first modify the video captions using a language model [16]. Then, we provide the edited captions along with the original videos to the video editing models, which generates the new videos.

**Editing Captions.** To generate our edited captions, we focus on 4 key visual factors that can be changed. While there are many possible factors, video captions help narrow this down by emphasizing important visual elements and ignoring unnecessary details [11]. In our approach, we break down video captions into a set of structured components : "Object Attribute", "Object Action", "Background", and "Style". We use these visual factors to edit the captions.

**Editing Videos.** To ensure high-quality edited videos, we utilize various State-of-the-Art (SoTA) editing models in our pipeline based on their performance in our evaluation experiments. Specifically, we propose four key evaluation metrics including *Fantasy Level (FL)*, *Scene Complexity (SC)*, *Camera motion (CM)*, and *Object Motion (OM)* and evaluate these editing models' performance. Subsequently, we select the top three models based on the experiment result for video editing. Finally, we conduct a rigorous manual filtering process to ensure the video quality. As a result, we generate high-quality 2.1k video-caption pairs.

| Model | LLM | Action Recognition | | | Captioning | | | Object Existence | | | Object Recognition | | | Overall | | |
|---|---|---|---|---|---|---|---|---|---|---|---|---|---|---|---|---|
| | | Origin | Edit | Drop | Origin | Edit | Drop | Origin | Edit | Drop | Origin | Edit | Drop | Origin | Edit | Drop |
| Larger or fine-tuned video encoder | | | | | | | | | | | | | | | | |
| VideoChat [1] | Vicuna-7B | 65.42 | 37.41 | -28.01 | 69.16 | 48.27 | -20.89 | 75.50 | 64.81 | -10.69 | 63.85 | 45.47 | -18.38 | 68.48 | 48.99 | -19.50 |
| VideoChat2 [2] | Mistral-7B | 67.08 | 51.41 | -15.67 | 74.29 | 62.60 | -11.69 | 75.02 | 72.19 | -2.83 | 68.27 | 57.13 | -11.14 | 71.17 | 60.83 | -10.34 |
| VideoLLaMA2 [3] | Mistral-7B | 65.15 | 41.11 | -24.04 | 76.55 | 59.25 | -17.30 | 77.83 | 69.73 | -8.10 | 62.17 | 47.31 | -14.86 | 70.43 | 54.35 | -16.08 |
| VideoLLaVA [4] | Vicuna-7B | 73.62 | 55.58 | -18.04 | 59.24 | 46.30 | -12.94 | 78.13 | 64.51 | -13.62 | 69.21 | 57.38 | -11.83 | 70.05 | 55.95 | -14.10 |
| VideoLLaMA3 [5] | Qwen2.5-7B | 75.11 | 60.30 | -14.81 | 78.49 | 64.09 | -14.40 | 82.47 | 72.78 | -9.69 | 71.24 | 59.20 | -12.04 | 76.83 | 64.09 | -12.73 |
| CLIP ViT/L-14 | | | | | | | | | | | | | | | | |
| VideoChatGPT [6] | Vicuna-7B | 55.37 | 29.33 | -26.04 | 65.29 | 41.17 | -24.12 | 75.74 | 57.15 | -18.59 | 67.64 | 45.75 | -21.89 | 66.01 | 43.35 | -22.66 |
| mPLUG-Owl3 [7] | Qwen2-7B | 69.26 | 43.72 | -25.54 | 63.85 | 42.71 | -21.14 | 73.88 | 63.13 | -10.75 | 69.09 | 52.17 | -16.92 | 69.02 | 50.46 | -18.56 |
| LLaVA-Next [8] | Vicuna-7B | 64.17 | 24.39 | -39.78 | 68.11 | 40.01 | -28.10 | 78.61 | 60.60 | -18.01 | 69.01 | 48.65 | -20.36 | 69.98 | 43.41 | -26.56 |
| LLaVA-Next$_{ori}$ | Vicuna-7B | 77.34 | 29.45 | -47.89 | 72.94 | 47.75 | -25.19 | 80.27 | 62.21 | -18.06 | 72.16 | 46.55 | -25.61 | 77.34 | 46.49 | -30.85 |
| LLaVA-Next$_{Ro}$ | Vicuna-7B | 86.21 | 81.36 | -4.85 | 73.12 | 69.68 | -3.44 | 81.13 | 77.58 | -3.55 | 75.76 | 68.29 | -7.47 | 79.05 | 74.23 | -4.83 |
| Overall | | 66.90 | 42.91 | -23.99 | 69.37 | 50.55 | -18.82 | 77.15 | 65.61 | -11.54 | 67.56 | 51.63 | -15.93 | 70.25 | 52.68 | -17.57 |

**Table 1**. Evaluation results on Ro-Bench for four tasks: 'Origin' (↑) denotes the test accuracy on the original video, 'Edit' (↑) represents the test accuracy on the edited video, and 'Drop' (↓) indicates the degradation in performance. The 'Drop' in model performance is highlighted in red. The highest robustness performance is marked with a red block, while the best performances of models using different encoders are indicated with blue and green blocks, respectively.

| Method | LLM | AS | AP | AA | FA | UA | OE | OI | OS | MD | AL | ST | AC | MC | MA | SC | FP | CO | EN | ER | CI | Avg. |
|---|---|---|---|---|---|---|---|---|---|---|---|---|---|---|---|---|---|---|---|---|---|---|
| GPT-4V [16] | - | 55.5 | 63.5 | 72.0 | 46.5 | 73.5 | 18.5 | 59.0 | 29.5 | 12.0 | 40.5 | 83.5 | 39.0 | 12.0 | 22.5 | 45.0 | 47.5 | 52.0 | 31.0 | 59.0 | 11.0 | 43.5 |
| Otter-V [17] | LLaMA-7B | 23.0 | 23.0 | 27.5 | 27.0 | 29.5 | 53.0 | 28.0 | 33.0 | 24.5 | 23.5 | 27.5 | 26.0 | 28.5 | 18.0 | 38.5 | 22.0 | 22.0 | 23.5 | 19.0 | 19.5 | 26.8 |
| mPLUG-OWL-V [18] | LLaMA-7B | 22.0 | 28.0 | 34.0 | 29.0 | 29.0 | 40.5 | 27.0 | 31.5 | 27.0 | 23.0 | 29.0 | 31.5 | 27.0 | 40.0 | 44.0 | 24.0 | 31.0 | 26.0 | 20.5 | 29.5 | 29.7 |
| Video-LLAMA [19] | Vicuna-7B | 27.5 | 25.5 | 51.0 | 29.0 | 39.0 | 48.0 | 40.5 | 38.0 | 22.5 | 22.5 | 43.0 | 34.0 | 22.5 | 32.5 | 45.5 | **32.5** | 40.0 | 30.0 | 21.0 | 37.0 | 34.1 |
| Video-ChatGPT [6] | Vicuna-7B | 23.5 | 26.0 | 62.0 | 22.5 | 26.5 | 54.0 | 28.0 | 40.0 | 23.0 | 20.0 | 31.0 | 30.5 | 25.5 | 39.5 | **48.5** | 29.0 | 33.0 | 29.5 | 26.0 | 35.5 | 32.7 |
| VideoChat [1] | Vicuna-7B | 33.5 | 26.5 | 56.0 | 33.5 | 40.5 | 53.0 | 40.5 | 30.0 | 25.5 | 27.0 | 48.5 | 35.0 | 20.5 | **42.5** | 46.0 | 26.5 | **41.0** | 23.5 | 23.5 | 36.0 | 35.5 |
| LLaVA-Next [8] | Vicuna-7B | 23.5 | 28.0 | 50.0 | 26.5 | 25.5 | 52.0 | 29.5 | 37.0 | 26.5 | 25.5 | 33.5 | 28.0 | 25.5 | 33.0 | 40.0 | 23.5 | 39.5 | 32.0 | 27.5 | 24.5 | 31.55 |
| LLaVA-Next$_{Ro}$ | Vicuna-7B | **41.5** | **48.5** | **70.5** | **55.0** | **49.5** | **66.5** | **52.5** | **44.0** | **29.0** | **33.5** | **51.0** | **41.0** | **37.0** | 42.0 | 44.0 | 30.5 | 39.0 | **41.0** | **30.0** | **40.5** | **44.33**[(+12.78)] |

**Table 2**. Results of MVBench, with the best performance on tasks highlighted in bold. The average improvement of LLaVA-Next$_{Ro}$ over LLaVA-Next is marked in red. Abbreviations like AS represent different tasks.

### 2.3. QA Pairs Generation

We carefully craft generation rules for both the questions and their corresponding answer options. The detailed procedure for generating QA pairs is outlined as follows.
**Automated Question Generation**: We leverage large language models to automate the conversion of video annotations into the desired format. Specifically, we first utilize GPT-4o [16] to generate a question for each video based on the corresponding task definition. Following this, we construct the answer options as outlined below.
**Automatic Option Generation:** (1) **Adopting from annotations:** For tasks such as Action Recognition", Object Existence", and Object Recognition", the correct options are directly derived from ground truth captions, with a Large Lan-guage Model (LLM) extracting the relevant information. (2) **LLM-based generation:** For object existence tasks, the options are yes," no," and not sure." For other tasks, distractors are designed to be neither trivial nor overly difficult. In action-related tasks, candidates are informed by both the action and its subject; for caption tasks, the original caption is split into components (attribute, object-action, background, style, subject), and candidates are created by altering them to ensure both relevance and diversity. Finally, options are shuffled to enhance robustness.

## 3. EVALUATION EXPERIMENTS

### 3.1. Robustness evaluation results

**The impact of editing factors on model performance**. As shown in Table 1, videos with counterfactual visual information cause significant performance drops across all models, revealing a vulnerability to rare and unseen concepts. This challenges their deployment in high-stakes applications. For instance, VideoChat2 [2] shows relative robustness (10.34% accuracy drop), while LLaVA-Next is more sensitive (26.56% drop).

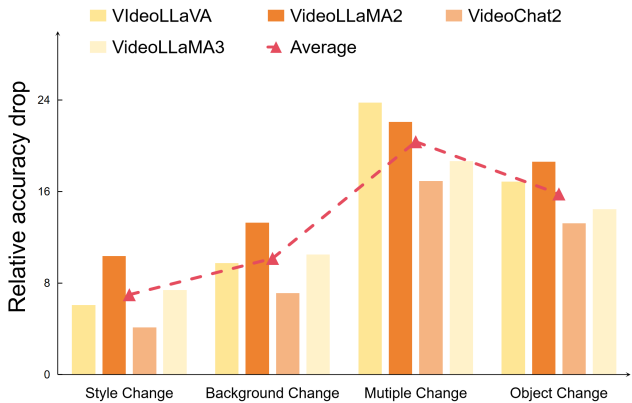

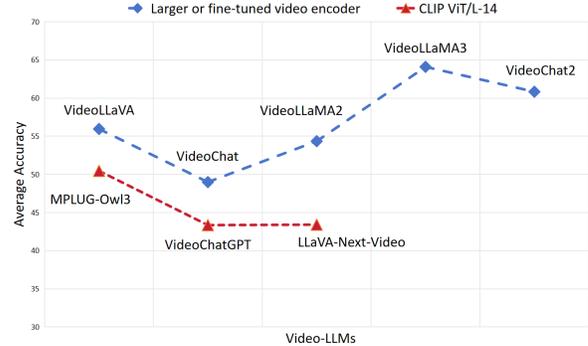

**Fig. 3**. Accuracy drop of different models across tasks. 'Average' represents the mean drop across models on various tasks.

**Fig. 4**. Accuracy comparison of MLLMs using CLIP ViT/L-14 (red) versus larger or fine-tuned video encoders (blue).

**The impact of editing factors on different tasks**. Visual variations most significantly impact action recognition tasks (23.99% performance drop) compared to object existence tasks (11.54% drop). We argue this is because disrupting dynamic visual cues, crucial for temporal reasoning, can cause model "hallucinations," highlighting the need for improved robustness in time-sensitive tasks.

**Comparison of the impact of different editing factors**. As depicted in Figure 3, models are more sensitive to object variations than to variations in style and background changes. For instance, the impact of background and style editing factors on model performance is nearly identical and significantly lower than the effect of object variations. We attribute this to local attributes (e.g., object appearance) being more critical for distinguishing key objects and actions than global attributes.

**The impact of model architecture on model performance**. As shown in Figure 4, models using the frozen CLIP ViT/L-14 as the video encoder perform worse than models that utilize larger or fine-tuned video encoders. This suggests that more powerful video encoders are crucial for exploring comprehensive video features and enhancing the robustness of multimodal video models.

### 3.2. Fine-Tuned Model Results

We used our pipeline to construct a training dataset. Our training dataset consists of 332 original videos, 1328 realistic counterfactual video samples, and 6640 QA pairs. We fine-tuned the LLaVA-Next model on our training set donated as **LLaVA-Next$_{Ro}$**. To demonstrate the effectiveness of our custom training data, we also train LLaVA-Next using unedited video data donated as **LLaVA-Next$_{ori}$**.

**Improvement of robustness on our benchmark**. As shown in Table 1, the LLaVA-Next$_{Ro}$ achieves the SoTA performance on our benchmark's robustness evaluation, surpasses VideoChat2 with an accuracy degradation of 4.83%. Furthermore, compared to the **LLaVA-Next**, there is a significant increase of 21.73% in the robustness evaluation metrics. Compared to **LLaVA-Next$_{ori}$**, LLaVA-Next$_{Ro}$ utilizes richer and more diverse data, enabling the model to maintain robust performance even when confronted with rare and uncommon counterfactual data. This demonstrates that our custom training data, constructed through our pipeline, significantly enhances the model's robustness. In addition to the improved robustness, the LLaVA-Next$_{Ro}$ demonstrates notable performance gains across all four downstream tasks.

**Improvement of video understanding capability**. To demonstrate the effectiveness of our custom training data, we tested our LLaVA-Next$_{Ro}$ on MVBench [2]. Table 2 presents our experiment results. LLaVA-Next$_{Ro}$ demonstrates consistent improvements across 20 downstream video understanding tasks, with an average performance gain of 12.78%. Notably, tasks related to actions and objects exhibit even more significant improvements. Notably, tasks related to actions and objects exhibit even more significant improvements. These experimental results highlight the importance of robustness evaluation using reliable counterfactual video test sets, and further prove that incorporating high-quality, reliable counterfactual perturbation data during fine-tuning enhances the model's video understanding capabilities.

### 4. CONCLUSION

In this work, we introduce Ro-Bench, a benchmark and training dataset created via an automated pipeline to evaluate and enhance MLLM robustness. Our evaluation on Ro-Bench reveals significant limitations in model robustness, but we demonstrate that fine-tuning with our dataset effectively enhances robustness. Overall, this work provides a systematic framework and valuable insights for advancing the development of more reliable and robust multimodal video models.